\documentclass[conference]{IEEEtran}
\IEEEoverridecommandlockouts
\usepackage[numbers]{natbib}
\usepackage{amsmath,amssymb,amsfonts}
\usepackage{algorithmic}
\usepackage{graphicx}
\usepackage{textcomp}
\usepackage{xcolor}
\usepackage{comment}
\usepackage{todonotes}
\usepackage{balance}
\def\BibTeX{{\rm B\kern-.05em{\sc i\kern-.025em b}\kern-.08em
    T\kern-.1667em\lower.7ex\hbox{E}\kern-.125emX}}

\begin{document}

\title{Prompt-to-OS (P2OS): Revolutionizing Operating Systems and Human-Computer Interaction with Integrated AI Generative Models\\
}

\newcommand{\linebreakand}{%
  \end{@IEEEauthorhalign}
  \hfill\mbox{}\par
  \mbox{}\hfill\begin{@IEEEauthorhalign}
}

\author{\IEEEauthorblockN{Gabriele Tolomei,\textsuperscript{\textsection} Cesare Campagnano,\textsuperscript{\textsection}}
\IEEEauthorblockA{\textit{Department of Computer Science} \\
\textit{Sapienza University of Rome, Italy}\\
\{tolomei, campagnano\}@di.uniroma1.it}
\and
\IEEEauthorblockN{Fabrizio Silvestri, Giovanni Trappolini}
\IEEEauthorblockA{\textit{Department of Computer Engineering} \\
\textit{Sapienza University of Rome, Italy}\\
\{fsilvestri, trappolini\}@diag.uniroma1.it}
}

\maketitle

\begingroup\renewcommand\thefootnote{\textsection}
\footnotetext{Equal contribution.}
\endgroup

\begin{abstract}

In this ambitious paper, we present a groundbreaking paradigm for human-computer interaction that revolutionizes the traditional notion of an operating system. 
Within this innovative framework, user requests issued to the machine are handled by an interconnected ecosystem of generative AI models that seamlessly integrate with or even replace traditional software applications. At the core of this paradigm shift are large generative models, such as language and diffusion models, which serve as the central interface between users and computers. This pioneering approach leverages the abilities of advanced language models, empowering users to engage in natural language conversations with their computing devices.

By capitalizing on the power of language models, users can articulate their intentions, tasks, and inquiries directly to the system, eliminating the need for explicit commands or complex navigation. The language model comprehends and interprets the user's prompts, generating and displaying contextual and meaningful responses that facilitate seamless and intuitive interactions.

This paradigm shift not only streamlines user interactions but also opens up new possibilities for personalized experiences. Generative models can adapt to individual preferences, learning from user input and continuously improving their understanding and response generation. Furthermore, it enables enhanced accessibility, as users can interact with the system using speech or text, accommodating diverse communication preferences.

However, this visionary concept also raises significant challenges, including privacy, security, trustability, and the ethical use of generative models. Robust safeguards must be in place to protect user data and prevent potential misuse or manipulation of the language model.

While the full realization of this paradigm is still far from being achieved, this paper serves as a starting point for envisioning the transformative potential of a human-computer interaction paradigm centered around artificial intelligence. We discuss the envisioned benefits, challenges, and implications, paving the way for future research and development in this exciting and promising direction. 
\end{abstract}

\begin{IEEEkeywords}
AI generative models for operating systems, AI generative models for human-computer interaction,
AI generative models as universal applications
\end{IEEEkeywords}

\section{Introduction}
The evolution of human-computer interaction (HCI) has undergone several transformations over the decades, with technology continuously striving to make computers more user-friendly and accessible. From the command-line interfaces of the 1960s to the graphical user interfaces (GUI) of the 1980s and, more recently, the touch interfaces on mobile devices, each shift has represented a significant leap towards more intuitive, efficient, and seamless user experiences. Today, as we find ourselves at the precipice of another paradigm shift, the question is not whether, but how, we continue to shape this ongoing evolution to ensure a future where technology serves us in increasingly human-centric ways.

In the current technological landscape, artificial intelligence (AI) stands as a powerhouse of potential, particularly for augmenting and redefining current operating systems and user interfaces. The abilities of large generative models (LGMs), such as large language models (LLMs) and diffusion models (DMs), have given us a glimpse into a future where our interactions with technology transcend the traditional boundaries. 

LLMs, built upon vast data sets and sophisticated architectures, are capable of completing complex tasks, demonstrating chain-of-thought reasoning akin to human capabilities, and displaying impressive generalization skills. Their proficiency in comprehending and generating language makes them ideal base-reasoners, capable of orchestrating diverse system components to create a seamless, intuitive, and responsive user interface.

Moreover, with advances in generative computer vision models, especially DMs, our toolbox for enhancing human-computer interaction has expanded. These models can generate incredibly realistic outputs, setting the stage for them to serve as the foundation for \textit{user interface generation}: the ability to generate personalized interfaces on-the-fly, that cater and adapt to individual user preferences, their character, and mood marks a shift toward highly customized and user-centric design, a shift that promises to enrich user experiences significantly.

This new paradigm of human-computer interaction presents exciting opportunities, such as enabling communication between systems that otherwise do not integrate the same API. By utilizing natural language, a universal medium, we can bridge the gap between disparate systems, fostering a more unified, coherent, and efficient interaction landscape.

However, this shift in paradigm also brings its share of challenges. A prime example is the need to ensure data persistence within these models. One key question when implementing this new approach is how we can keep a consistent and ongoing dialogue over time, especially when the system is working on complicated or multi-stage tasks. This steady interaction is crucial for a smooth user experience and for building trust in the system's ability to assist the user effectively. To make this possible, we may need to step away from the methods we're used to and start thinking about new ways to improve the performance of these generative models. For instance, current methods of data management, such as storing files explicitly in computers or data centers, may provide some benefits, but they may not fully meet the unique needs of generative models, which store their knowledge implicitly, compressed within their parameters.

While the capabilities of LLMs in understanding and generating language are remarkable, they are not without their limitations. 
These issues primarily originate from the data employed for their pre-training, which is frequently obtained from web crawls. This data can often contain biased, toxic, or harmful content, consequently impairing models' reliability. Another limitation is the tendency to hallucinate, i.e., despite not having any explicit misinformation, LLMs may generate outputs that are not entirely accurate or faithful. This propensity to deviate from the input can occasionally lead to responses that, while contextually plausible, might misrepresent the user's intent or the factual information at hand.

Moreover, the promise of seamless interaction and communication must balance with considerations of trustability, privacy, security, and ethics. For this reason, developing new protocols for information exchange becomes a necessity in this envisioned future. These protocols must meet and surpass current standards, protecting user data while simultaneously ensuring private and secure interactions. The design of such protocols also must anticipate and be resilient against potential misuse of AI systems, providing robust safeguards to exploitation and unethical practices.

These represent just a few of the challenges in harnessing the full potential of LLMs in revolutionizing human-computer interaction. As we venture into this exciting new territory, it is essential to confront these challenges head-on, ensuring that the solutions we develop are not just technologically advanced, but also reliable, ethical, and user-centric.

The road ahead in this new paradigm is both promising and challenging. This paper serves as an exploration into the future of human-computer interaction – a future where our interactions with technology become akin to a natural conversation. We delve deeper into the benefits, challenges, and implications of this envisioned future in the following sections, charting a course for continued research and development in this transformative and exciting direction. In particular, Section~\ref{sec:background} reviews current work in this area, and provides an idea of the current technological landscape; Sections~\ref{sec:vision} and~\ref{sec:proposed-solution} describe our vision and propose a possible architecture, respectively. Section~\ref{sec:challenges} questions the main challenges that may arise; finally, Section~\ref{sec:conclusion} concludes our discussion.


\section{Related Work}\label{sec:background}

Recent years have seen the rise of \textit{Transformers} as the leading architecture for (deep) learning systems.
Initially introduced as a technique for machine translation \cite{vaswani2017attention}, they have soon been recognized as valuable for text-related downstream tasks.
Works such as \cite{devlin2018bert,clark2020electra,liu2019roberta,campagnano2022srl4e} pre-train a transformer on a self-supervised task and finetune it on a specific downstream task, usually using a small amount of data and achieving super-human performance \cite{wang2018glue}. 
Others \cite{brown2020language,radford2018improving,radford2019language} have focused on using transformers as generative LLMs.
These models effectively train to predict the next token sequence given a particular context/input.
This latter strategy has reached wide popularity, even among the general public, thanks to recent successes like \cite{openai-chatgpt}.
These LLMs have fully exploited the transformer's capacity to scale to a huge number of parameters, allowing them to have exceptional capabilities on many downstream tasks, even in a zero-shot setting, i.e., without requiring further supervision on the specific task.
Even more surprisingly, the models can improve these tasks using \emph{prompting} \cite{lu2021fantastically}.
Prompting consists in providing specific input to the LLM, inducing a more accurate response by the model.
We find other techniques inside this paradigm, like that of \emph{in-context learning} \cite{dong2023survey}, that is, to provide the LLM with a few examples of the task in its input, sometimes greatly enhancing its performance.
We would like the reader to notice that these techniques do not require any additional training and can be performed at inference time.
While these LLMs have shown impressive capabilities, they also have several shortcomings.
Above all, the inability to deal with a large context/input \cite{liu2023lost} and the tendency to hallucinate \cite{zhang2023language} has led researchers to look for ways of augmenting them \cite{mialon2023augmented,mmndb,thorne-etal-2021-database}.
Among these, we find most interesting for this work the line of researchers that aims at augmenting LLMs with the use of tools \cite{schick2023toolformer,patil2023gorilla}.
Under this paradigm, the LLM can call for help in the form of APIs.
For instance, the model could call a calculator to perform a mathematical operation.

In this paper, we want to move beyond this paradigm.
While the LLM can call for tools, they are imagined as strictly rigid and static APIs.
What we envision, instead, is to have generative models that can communicate both with the user and among each other in a natural manner; in other words, we foresee the \emph{the end of programming} \cite{endofp} as we know it for the average computer user.

This new proposed paradigm opens up a realm of possibility.
For example, thanks to recent advances in text-to-image generation \cite{rombach2022highresolution}, authors from \cite{wei2023boosting} have proposed a GUI generated at runtime and explicitly personalized for a particular user-task-experience triplet.
Even further, deep learning models are now reaching multimodal capabilities beyond just images, with methods proficient on audio \cite{dhariwal2020jukebox,borsos2023audiolm,barnabo2023cycledrums}, video \cite{VideoFusion,li2022mvitv2}, and 3D \cite{chen2023scenedreamer,trappolini2021shape,halimi2020towards}.
In the following sections, we develop more thoroughly this vision that promises to revolutionize system design and human-computer interactions.

\section{Vision}\label{sec:vision}
Let us consider a hypothetical scenario as an exemplification to introduce this section. 
In the following, we will use the term ``\textit{agent}'' to indicate any generative AI model within our ecosystem.
We denote by $A_c$ a client agent, which is a user's personal assistant, while $A_s$ designates a set of server agents, which provide specific services or resources.

We illustrate this through the following dialogue sequence, where \textit{files} and \textit{actions} are represented by square brackets:

\begin{enumerate}
\item User to $A_c$: ``Please find me a flight to Paris on the 16th or 17th of July in the evening for less than 120 USD. Don't show me all the options; propose the cheapest one directly''
\item $A_c$ to $[A_{s1}, A_{s2}, A_{s3}]$: ``My user would like a flight to Paris, between the 16th and 17th of July, preferably in the evening, at a cost not exceeding 120 USD. Who can provide options?''
\item $A_{s1}$ to $A_c$: ``No options available.''\\
$A_{s2}$ to $A_c$: ``There are two flights, the first is on July 16th at 4 PM for 118 USD, the other on July 17th at 6 PM for 95 USD.''\\
$A_{s3}$ to $A_c$: ``There is a flight on July 17th at 4 PM for 110 USD.''
\item $A_c$ to $A_{s2}$: ``Please send a quote for the second option (the one on July 17th at 6 PM for 95 USD).''
\item $A_{s2}$ to $A_c$: [PDF][Secure payment link]
\item $A_c$ to User: ``Here is the quote for a 95 USD flight [Display PDF]. Do you want to book?''
\item User to $A_c$: ``Yes, and then print the ticket.''
\item $A_c$ to $A_{s2}$: [Payment]
\item $A_{s2}$ to $A_c$: [Ticket PDF]
\item $A_c$ to User: ``Here is the ticket. [Display ticket PDF][Print ticket PDF]''
\end{enumerate}


This hypothetical scenario above is just one example of a broader vision that is mappable to this framework. This system could have enormous flexibility, for example, ``find me a flight to go to Thailand, Indonesia, Vietnam, or Cambodia around mid-August for 14-16 days. When you find something less than 700 USD, book it without asking for my confirmation, and print me the ticket''.

In this scenario, $A_c$ serves as the \textit{orchestrator}, coordinating with the server agents $A_s$ to achieve the user's goal. $A_c$, empowered by LLMs and DMs, is capable of understanding the user's instructions, delegating tasks, and managing the dialogue's flow. Meanwhile, the server agents $A_s$, powered by various specialized components/models, handle specific tasks such as finding flights, handling payments, and generating PDFs.

To further delineate our vision, let us delve deeper into the integration of state-of-the-art computer vision systems and their potential role in on-the-fly GUI rendering. Diffusion models, which have demonstrated remarkable capabilities in generating high-quality, realistic outputs, could serve as a cornerstone for this task.

One of the exciting prospects that the adoption of such models brings is the ability to customize and personalize the interface in response to the user's preferences, character, and even mood. Given that these models can be trained to generate a wide range of visual outputs, they could be directed to design interfaces that echo a user's aesthetic preferences or adapt to their current mood. For instance, the system could switch from a minimalist design with soft colors to a vibrant, dynamic design as it detects a change in the user's emotional state.

Such a degree of customization would revolutionize the concept of user-centric design, moving away from static designs to more fluid and responsive ones. This ability to generate personalized interfaces on-the-fly represents a significant shift towards a future where technology can deeply integrate into our lives, responding to our needs and moods in real time.

Additionally, the integration of speech-to-text and text-to-speech models can further enhance this future vision of human-computer interaction. The coupling of these models with a powerful language understanding system allows for interactions that are more in line with natural, human conversation. Users could convey their needs verbally, and the system could respond in kind, further blurring the lines between human-computer interaction and human-human conversation.

For instance, in the previously discussed scenario, the user could verbalize their request for a flight booking, and the client agent ($A_c$) could acknowledge, confirm, and execute these instructions using spoken language. This seamless integration of speech-to-text and text-to-speech models would provide an interaction experience that is not just intuitive but also highly efficient, especially for users with visual impairments or those who are occupied with other tasks and prefer to interact verbally with their devices.

Furthermore, these systems could extend beyond serving individual users and facilitate interactions between groups of users. For instance, they could be deployed in conference calls or group meetings, transcribing the conversation, summarizing the key points, and even responding to queries in real time. This transition to more natural and fluid forms of communication holds immense promise for both personal and professional contexts, signaling a future where our interaction with technology is as natural and intuitive as speaking with a friend or colleague.

This concept of personalized, responsive, and accessible interfaces could even extend to other sensory modalities, such as haptics, further broadening the scope of HCI. With advancements in AI and ML, the future of HCI could encompass an array of sensory interactions, each tailored to individual user needs and preferences, creating an immersive and inclusive technological environment.

Despite these potential advancements, it is important to keep in mind that such a high level of customization and personalization carries with it a host of challenges related to privacy, security, and ethics. Addressing these challenges is essential to ensure the successful implementation of this vision. These challenges and potential solutions are discussed in greater detail in Sections~\ref{sec:challenges} and~\ref{sec:proposed-solution}.

\section{Proposed Architecture}\label{sec:proposed-solution}
As the trajectory of human-computer interaction continues to evolve, we propose a novel architecture that shifts the paradigm of traditional system design. This architecture is visualized in Figure 1, highlighting the integration of an LLM directly above the system call layer of the operating system, fundamentally altering how users (and current standalone applications) interface with low-level computing resources.

\begin{figure}[h!]
\centering
\includegraphics[width=0.5\textwidth]{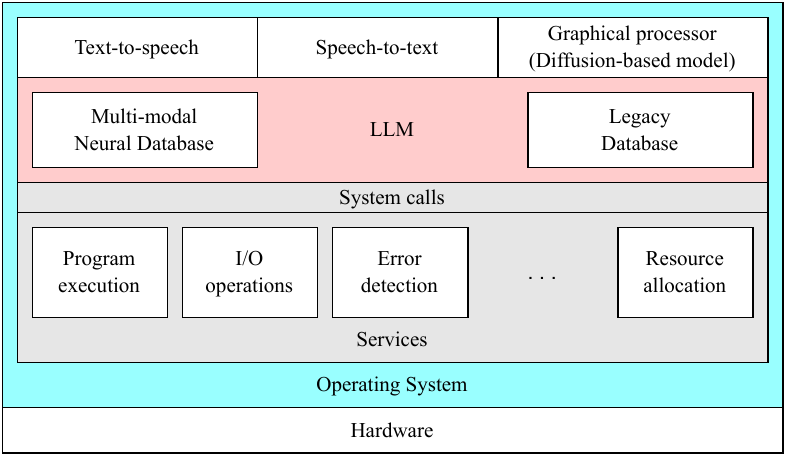}
\caption{A representation of the proposed architecture highlighting the operating system, system calls, LLM, and its integration with a multi-modal neural database and other integrated components.}
\label{fig:architecture}
\end{figure}

\subsection{High-Level to Direct LLM Interface}
Referring to Figure~\ref{fig:architecture}, traditionally, user requests, translated by standalone applications, would interface with the operating system through \textit{system calls}. Our proposed architecture envisions an LLM layer sitting atop the system call layer. This LLM would harness the power of system calls to communicate with the operating system directly. By introducing this intermediary LLM layer, we can achieve several advantages:

\begin{itemize}
\item \textbf{Reduction in Redundancy:} Redundancies inherent in maintaining multiple application layers, especially those with overlapping backend processes, can be significantly reduced.
\item \textbf{Simplified Communication:} LLMs, acting as universal mediators, can execute user commands across various platforms, obviating the need for shared APIs.
\item \textbf{Natural Interaction:} Users can employ natural language, moving away from domain-specific commands, promoting a more human-centric design.
\end{itemize}

\subsection{Transitioning Away from Standalone Applications}
While the LLM becomes the primary interface, standalone applications will not vanish but transform. They could serve as specialized plug-ins or tools for the LLM. For complex tasks, such as advanced graphic design, specialized applications might still be employed. However, initiation and basic interactions can be handled in natural language, seamlessly integrating these applications with the LLM.

\subsection{Integrated Components}
In addition to the LLM, our architecture encompasses:

\begin{itemize}
\item \textbf{Graphical Processor based on Diffusion Models:} This caters to visual tasks, allowing for the generation and interpretation of personalized user interfaces and graphical content, ensuring a multi-modal interaction platform.
\item \textbf{Multi-modal Neural Database:} Integral to our proposed architecture, as visualized in Figure~\ref{fig:architecture}, is the connection of the LLM to a multi-modal neural database. This database serves as persistent memory storage for the LLM, ensuring consistent user experiences across sessions. Unlike traditional databases that store explicit data, this neural database retains information in a format amenable to direct neural processing, facilitating immediate and efficient data retrieval and modification by the LLM.
\item \textbf{Text-to-Speech and Speech-to-Text Systems:} These components allow for auditory interactions, where users can speak to and receive vocal feedback from the system.
\end{itemize}
The outlined architecture supports adaptability and user-centricity, with components continuously refining their operations based on feedback. Data security, consistency, and the reliability of LLM-mediated interactions remain pivotal, and will be discussed in more detail in the next section.

\section{Challenges}\label{sec:challenges}
Developing an operating system that integrates generative AI models like the one sketched in Figure~\ref{fig:architecture} promises to reshape system design dramatically. Therefore, realizing this vision is not without its challenges, which span technological, security, privacy, and ethical domains. In this section, we delve deeper into these obstacles, highlighting the complex and multifaceted work required to bring this vision to fruition.

\subsection{Trustability and Safety}
LLMs have shown remarkable capabilities in understanding and generating human-like text. This remarkable prowess stems from their training on enormous, diverse web data, which allows them to assimilate an impressive understanding of language structure, context, and information.

However, this training approach can also present substantial risks \cite{zugner2020adversarial,chakraborty2018adversarial,trappolini2022sparse}. Uncontrolled web data, which is often part of large pre-training corpora, can introduce the possibility of biased, harmful, or toxic behavior. For instance, if the training data includes prejudiced viewpoints or false information, the LLM might inadvertently absorb these biases and misinformation, potentially affecting the quality and trustworthiness of the content it generates.

Moreover, LLMs can sometimes generate plausible yet inaccurate content, a phenomenon often referred to as hallucination. Users may overlook minor inaccuracies in the short term, but persistent or significant misrepresentations can erode trust in the system. This challenge becomes even more acute when the system handles sensitive or critical tasks, where accuracy and reliability are paramount.

Developing methods to reduce these hallucinations, increase the trustworthiness of generated content, and provide transparency into the system's decision-making process is a significant, non-trivial challenge in realizing this vision.

\subsection{Technological Challenges}
\subsubsection{Data Persistence}
Current LLMs are stateless, meaning they do not maintain a memory of past interactions. While this is not an issue for single, isolated tasks, it poses a substantial challenge for complex, multi-stage tasks that require an ongoing dialogue with the user. In these scenarios, maintaining a consistent ``conversation thread'' is crucial for the system to function effectively and provide a smooth, seamless user experience.

More generally, LGMs encode their knowledge implicitly within their model parameters, effectively compressing vast amounts of information into a highly condensed form. While this approach allows the model to generate rich, contextually-aware content, it also presents a significant challenge in retrieving and using this information effectively.

Innovative solutions for data persistence and memory management are essential for achieving smooth, natural human-computer interactions. These solutions could involve novel data storage and retrieval mechanisms, new ways of representing and tracking dialogue states, or creative uses of meta-learning to adapt and personalize the system over time.

\subsubsection{Hardware Considerations}
The integration of LGMs into operating systems entails careful consideration of both low-level and high-level hardware aspects. At the low level, the system's design must efficiently manage the available hardware resources. Running LGMs in real-time could demand substantial computational resources, potentially stretching the limits of current systems. Innovative technologies to build small and specialized models for this new kind of operating system should be devised, as we cannot make any use of an LGM that will exhaust the system resources available.

At a high level, the envisioned system would necessitate a radical shift in software-hardware interaction. The generated user interfaces must work seamlessly across diverse hardware configurations, demanding extensive adaptability and compatibility. Ensuring the system's efficacy and efficiency across a wide range of hardware poses a significant technical challenge.

\subsection{Security and Privacy}
\subsubsection{Communication Security}
As we move towards a new paradigm of human-computer interaction, secure communication becomes even more critical. Current communication protocols, like certificate-based authentication, provide robust mechanisms for ensuring that an agent is legitimate before starting communication. However, as we transition to a system centered around LLMs, these protocols will need to be augmented or replaced with new techniques tailored to the unique challenges and opportunities of this paradigm.

\subsubsection{AI-Social Engineering}
The sophistication of LGMs can potentially be exploited in AI-social engineering attacks. In such scenarios, a malicious user might attempt to deceive the AI system into revealing sensitive information or execute harmful actions. These attacks can take many forms and can be challenging to anticipate or prevent, given the diverse and unpredictable nature of human interaction.

Preventing these attacks will require careful system design, including setting strict parameters on the system's behavior, determining which data can be shared and under what conditions, and designing mechanisms for user consent and control over the system's actions.

\subsection{Ethics}

The integration of LGMs into operating systems also raises numerous ethical concerns. One significant concern is their potential to displace human workers in specific professions. Such displacement could lead to widespread job losses, contributing to economic inequalities and causing societal disruption on a potentially large scale.

Moreover, the power of these technologies to synthesize and manipulate data poses unique risks. For example, LGMs can be harnessed to create deepfakes and fake news, which can then be used for nefarious purposes, including disinformation campaigns or identity theft. These models can generate malicious content, such as automated phishing emails or hate speech, exacerbating existing social and ethical dilemmas.

These ethical issues highlight the importance of establishing robust safeguards, regulations, and guidelines to prevent misuse and manage the societal impact of these technologies.

\section{Conclusion}\label{sec:conclusion}
The evolution of human-computer interaction, enhanced by the capabilities of LGMs such as LLMs and DMs, has the potential to reshape system design and the dynamics of communication, interaction, and collaboration between users and machines. Through the integration of AI into operating systems, we envision a future where interfaces are not only intuitive but also deeply personalized, adapting to individual needs and preferences, allowing for seamless and coherent interactions. This paper offers a glimpse of such a transformative future, emphasizing both its unprecedented benefits and its multifaceted challenges.

Yet, the roadmap to such a future is not without its intricacies. Harnessing the full potential of AI-enhanced human-computer interaction requires navigating a landscape replete with challenges, ranging from data persistence, model reliability, bias, to the paramount concerns of trust, privacy, and ethical considerations. Addressing these challenges is not merely a technical exercise but a broader call for interdisciplinary collaboration. The complexity of these issues suggests that our current understanding and strategies may only be scratching the surface, necessitating a paradigm shift in our approach.

Despite these hurdles, this journey is of primary importance. The convergence of AI and HCI within system design can lead to a profound enhancement in the quality of our digital experiences, shifting from transactional commands to natural, conversation-like engagements. We envision a world where technology is a responsive collaborator, attuned to human needs and preferences. While this paper has just scratched the surface of this promising frontier, we hope that it serves as a catalyst, inspiring and guiding future research endeavors. 
\balance
\section*{Acknowledgment}
This work was partially supported by:
\begin{itemize}
    \item DRONES AS A SERVICE for FIRST EMERGENCY RESPONSE Project (Ateneo 2021);
    \item projects FAIR (PE0000013), SERICS (PE00000014), and  IR0000013-SoBigData.it under the MUR National Recovery and Resilience Plan funded by the European Union NextGenerationEU.
\end{itemize}

\bibliographystyle{ieeetr}
\bibliography{bib}

\begin{thebibliography}{10}

\bibitem{vaswani2017attention}
A.~Vaswani, N.~Shazeer, N.~Parmar, J.~Uszkoreit, L.~Jones, A.~N. Gomez,
  {\L}.~Kaiser, and I.~Polosukhin, ``Attention is all you need,'' {\em Advances
  in neural information processing systems}, vol.~30, 2017.

\bibitem{devlin2018bert}
J.~Devlin, M.-W. Chang, K.~Lee, and K.~Toutanova, ``Bert: Pre-training of deep
  bidirectional transformers for language understanding,'' {\em arXiv preprint
  arXiv:1810.04805}, 2018.

\bibitem{clark2020electra}
K.~Clark, M.-T. Luong, Q.~V. Le, and C.~D. Manning, ``Electra: Pre-training
  text encoders as discriminators rather than generators,'' {\em arXiv preprint
  arXiv:2003.10555}, 2020.

\bibitem{liu2019roberta}
Y.~Liu, M.~Ott, N.~Goyal, J.~Du, M.~Joshi, D.~Chen, O.~Levy, M.~Lewis,
  L.~Zettlemoyer, and V.~Stoyanov, ``Roberta: A robustly optimized bert
  pretraining approach,'' {\em arXiv preprint arXiv:1907.11692}, 2019.

\bibitem{campagnano2022srl4e}
C.~Campagnano, S.~Conia, and R.~Navigli, ``{SRL4E} {--} {S}emantic {R}ole
  {L}abeling for {E}motions: {A} unified evaluation framework,'' in {\em
  Proceedings of the 60th Annual Meeting of the Association for Computational
  Linguistics (Volume 1: Long Papers)}, pp.~4586--4601, 2022.

\bibitem{wang2018glue}
A.~Wang, A.~Singh, J.~Michael, F.~Hill, O.~Levy, and S.~R. Bowman, ``Glue: A
  multi-task benchmark and analysis platform for natural language
  understanding,'' {\em arXiv preprint arXiv:1804.07461}, 2018.

\bibitem{brown2020language}
T.~B. Brown, B.~Mann, N.~Ryder, M.~Subbiah, J.~Kaplan, P.~Dhariwal,
  A.~Neelakantan, P.~Shyam, G.~Sastry, A.~Askell, S.~Agarwal, A.~Herbert-Voss,
  G.~Krueger, T.~Henighan, R.~Child, A.~Ramesh, D.~M. Ziegler, J.~Wu,
  C.~Winter, C.~Hesse, M.~Chen, E.~Sigler, M.~Litwin, S.~Gray, B.~Chess,
  J.~Clark, C.~Berner, S.~McCandlish, A.~Radford, I.~Sutskever, and D.~Amodei,
  ``Language models are few-shot learners,'' 2020.

\bibitem{radford2018improving}
A.~Radford, K.~Narasimhan, T.~Salimans, and I.~Sutskever, ``Improving language
  understanding by generative pretraining,'' 2018.

\bibitem{radford2019language}
A.~Radford, J.~Wu, R.~Child, D.~Luan, D.~Amodei, and I.~Sutskever, ``Language
  models are unsupervised multitask learners,'' {\em OpenAI Blog}, 2019.

\bibitem{openai-chatgpt}
OpenAI, ``Chatgpt by openai,'' 2021.

\bibitem{lu2021fantastically}
Y.~Lu, M.~Bartolo, A.~Moore, S.~Riedel, and P.~Stenetorp, ``Fantastically
  ordered prompts and where to find them: Overcoming few-shot prompt order
  sensitivity,'' {\em arXiv preprint arXiv:2104.08786}, 2021.

\bibitem{dong2023survey}
Q.~Dong, L.~Li, D.~Dai, C.~Zheng, Z.~Wu, B.~Chang, X.~Sun, J.~Xu, L.~Li, and
  Z.~Sui, ``A survey on in-context learning,'' 2023.

\bibitem{liu2023lost}
N.~F. Liu, K.~Lin, J.~Hewitt, A.~Paranjape, M.~Bevilacqua, F.~Petroni, and
  P.~Liang, ``Lost in the middle: How language models use long contexts,'' {\em
  arXiv preprint arXiv:2307.03172}, 2023.

\bibitem{zhang2023language}
M.~Zhang, O.~Press, W.~Merrill, A.~Liu, and N.~A. Smith, ``How language model
  hallucinations can snowball,'' 2023.

\bibitem{mialon2023augmented}
G.~Mialon, R.~Dess{\`\i}, M.~Lomeli, C.~Nalmpantis, R.~Pasunuru, R.~Raileanu,
  B.~Rozi{\`e}re, T.~Schick, J.~Dwivedi-Yu, A.~Celikyilmaz, {\em et~al.},
  ``Augmented language models: a survey,'' {\em arXiv preprint
  arXiv:2302.07842}, 2023.

\bibitem{mmndb}
G.~Trappolini, A.~Santilli, E.~Rodol\`{a}, A.~Halevy, and F.~Silvestri,
  ``Multimodal neural databases,'' in {\em Proceedings of the 46th
  International ACM SIGIR Conference on Research and Development in Information
  Retrieval}, SIGIR '23, (New York, NY, USA), p.~2619–2628, Association for
  Computing Machinery, 2023.

\bibitem{thorne-etal-2021-database}
J.~Thorne, M.~Yazdani, M.~Saeidi, F.~Silvestri, S.~Riedel, and A.~Halevy,
  ``Database reasoning over text,'' in {\em Proceedings of the 59th Annual
  Meeting of the Association for Computational Linguistics and the 11th
  International Joint Conference on Natural Language Processing (Volume 1: Long
  Papers)}, (Online), pp.~3091--3104, Association for Computational
  Linguistics, Aug. 2021.

\bibitem{schick2023toolformer}
T.~Schick, J.~Dwivedi-Yu, R.~Dess{\`\i}, R.~Raileanu, M.~Lomeli,
  L.~Zettlemoyer, N.~Cancedda, and T.~Scialom, ``Toolformer: Language models
  can teach themselves to use tools,'' {\em arXiv preprint arXiv:2302.04761},
  2023.

\bibitem{patil2023gorilla}
S.~G. Patil, T.~Zhang, X.~Wang, and J.~E. Gonzalez, ``Gorilla: Large language
  model connected with massive apis,'' 2023.

\bibitem{endofp}
M.~Welsh, ``The end of programming,'' {\em Commun. ACM}, vol.~66, p.~34–35,
  dec 2022.

\bibitem{rombach2022highresolution}
R.~Rombach, A.~Blattmann, D.~Lorenz, P.~Esser, and B.~Ommer, ``High-resolution
  image synthesis with latent diffusion models,'' 2022.

\bibitem{wei2023boosting}
J.~Wei, A.-L. Courbis, T.~Lambolais, B.~Xu, P.~L. Bernard, and G.~Dray,
  ``Boosting gui prototyping with diffusion models,'' 2023.

\bibitem{dhariwal2020jukebox}
P.~Dhariwal, H.~Jun, C.~Payne, J.~W. Kim, A.~Radford, and I.~Sutskever,
  ``Jukebox: A generative model for music,'' {\em arXiv preprint
  arXiv:2005.00341}, 2020.

\bibitem{borsos2023audiolm}
Z.~Borsos, R.~Marinier, D.~Vincent, E.~Kharitonov, O.~Pietquin, M.~Sharifi,
  D.~Roblek, O.~Teboul, D.~Grangier, M.~Tagliasacchi, {\em et~al.}, ``Audiolm:
  a language modeling approach to audio generation,'' {\em IEEE/ACM
  Transactions on Audio, Speech, and Language Processing}, 2023.

\bibitem{barnabo2023cycledrums}
G.~Barnab{\`o}, G.~Trappolini, L.~Lastilla, C.~Campagnano, A.~Fan, F.~Petroni,
  and F.~Silvestri, ``Cycledrums: automatic drum arrangement for bass lines
  using cyclegan,'' {\em Discover Artificial Intelligence}, vol.~3, no.~1,
  p.~4, 2023.

\bibitem{VideoFusion}
Z.~Luo, D.~Chen, Y.~Zhang, Y.~Huang, L.~Wang, Y.~Shen, D.~Zhao, J.~Zhou, and
  T.~Tan, ``Videofusion: Decomposed diffusion models for high-quality video
  generation,'' in {\em Proceedings of the IEEE/CVF Conference on Computer
  Vision and Pattern Recognition (CVPR)}, June 2023.

\bibitem{li2022mvitv2}
Y.~Li, C.-Y. Wu, H.~Fan, K.~Mangalam, B.~Xiong, J.~Malik, and C.~Feichtenhofer,
  ``Mvitv2: Improved multiscale vision transformers for classification and
  detection,'' 2022.

\bibitem{chen2023scenedreamer}
Z.~Chen, G.~Wang, and Z.~Liu, ``Scenedreamer: Unbounded 3d scene generation
  from 2d image collections,'' 2023.

\bibitem{trappolini2021shape}
G.~Trappolini, L.~Cosmo, L.~Moschella, R.~Marin, S.~Melzi, and E.~Rodol{\`a},
  ``Shape registration in the time of transformers,'' {\em Advances in Neural
  Information Processing Systems}, vol.~34, pp.~5731--5744, 2021.

\bibitem{halimi2020towards}
O.~Halimi, I.~Imanuel, O.~Litany, G.~Trappolini, E.~Rodol{\`a}, L.~Guibas, and
  R.~Kimmel, ``Towards precise completion of deformable shapes,'' in {\em
  Computer Vision--ECCV 2020: 16th European Conference, Glasgow, UK, August
  23--28, 2020, Proceedings, Part XXIV 16}, pp.~359--377, Springer, 2020.

\bibitem{zugner2020adversarial}
D.~Z{\"u}gner, O.~Borchert, A.~Akbarnejad, and S.~G{\"u}nnemann, ``Adversarial
  attacks on graph neural networks: Perturbations and their patterns,'' {\em
  ACM Transactions on Knowledge Discovery from Data (TKDD)}, vol.~14, no.~5,
  pp.~1--31, 2020.

\bibitem{chakraborty2018adversarial}
A.~Chakraborty, M.~Alam, V.~Dey, A.~Chattopadhyay, and D.~Mukhopadhyay,
  ``Adversarial attacks and defences: A survey,'' {\em arXiv preprint
  arXiv:1810.00069}, 2018.

\bibitem{trappolini2022sparse}
G.~Trappolini, V.~Maiorca, S.~Severino, E.~Rodol{\`a}, F.~Silvestri, and
  G.~Tolomei, ``Sparse vicious attacks on graph neural networks,'' {\em arXiv
  preprint arXiv:2209.09688}, 2022.

\end{thebibliography}


\end{document}